\documentclass[letterpaper, 10 pt, conference]{ieeeconf}  

\IEEEoverridecommandlockouts                              

\usepackage{times}

\usepackage{svg}
\usepackage{amsmath}
\usepackage{graphicx}
\usepackage{amssymb}
\usepackage{mathtools}
\usepackage{subfig} 
\usepackage{adjustbox}
\usepackage{multirow}
\usepackage{float}
\usepackage{placeins}
\usepackage{hyperref}

\usepackage{tikz}
\usetikzlibrary{shapes.geometric, arrows}
\usetikzlibrary{3d,fit, positioning}

\usepackage{import}
\usetikzlibrary{quotes,arrows.meta}
\usetikzlibrary{positioning}

\usepackage{Box}
\usepackage{RightBandedBox}

\title{\LARGE \bf Content Disentanglement for Semantically Consistent Synthetic-to-Real Domain Adaptation}

\author{Mert Keser$^{1,2*}$\\TUM, BMW \and Artem Savkin$^{1,2*}$\\TUM, BMW \and Federico Tombari$^{1,3}$\\TUM, Google
\thanks{*Equal Contribution}
\thanks{$^{1}$TUM, 85748 Munich, Germany
{
\tt\small forename.surname@tum.de};
}%
\thanks{$^{2}$BMW AG, 80809 Munich (Germany)}
\thanks{$^{3}$Google, 8002 Zurich, Switzerland}}

\begin{document}

\maketitle

\begin{abstract}
Synthetic data generation is an appealing approach to generate novel traffic scenarios in autonomous driving. However, deep learning perception algorithms trained solely on synthetic data encounter serious performance drops when they are tested on real data. Such performance drops are commonly attributed to the domain gap between real and synthetic data. Domain adaptation methods that have been applied to mitigate the aforementioned domain gap achieve visually appealing results, but usually introduce semantic inconsistencies into the translated samples. In this work, we propose a novel, unsupervised, end-to-end domain adaptation network architecture that enables semantically consistent \textit{sim2real} image transfer. Our method performs content disentanglement by employing shared content encoder and fixed style code.
\end{abstract}

\section{Introduction}

Autonomous vehicles employ machine learning techniques in order to understand surrounding environments. This requires high generalization performance of an autonomous vehicle’s perception subsystem regarding the environments and traffic scenarios it might encounter in the real world. These environments and traffic scenarios include different variations of environmental factors, such as lighting and weather conditions, etc., near-accident scenarios, and so-called \textit{long tail} of events' distribution. Collecting sufficient training data that can cover various kinds of traffic scenarios in real-world environments is often not feasible. Moreover, such training data acquisition typically implies manual annotation. This might be a very laborious and time-consuming task \cite{cordts2016cityscapes}, especially in use-cases like semantic segmentation, as this process requires per-pixel labeling.

Synthetic data generation is a promising approach to overcome the described problem of training data acquisition. It is a cost-effective method where the annotation of the data can be generated practically at no cost and variance of the generated traffic scenarios potentially has no limit. However, the recognition algorithms trained with synthetic data reveals a significant performance accuracy drop when evaluation is done on real data \cite{zhao2020review}.

\input{figure_teaser}
\input{figure_architecture}

The performance drop is commonly attributed to domain gap between synthetic and real data distributions. Researchers tackle the gap problem by means of domain adaptation methods which typically rely on generative networks based on adversarial training in order to translate the synthetic data into a closer representation of the real data. These methods achieve visually appealing results, but the translation often includes semantic inconsistencies \cite{toldo2020unsupervised}. The primary approach for producing semantically consistent samples leverages existing knowledge about the scene in the form of semantic maps \cite{hoffman2018cycada, chen2019crdoco, li2019bidirectional}. They, however, rely on real data annotation, which we try to avoid in the first place.

In this work, we propose a new, unsupervised, end-to-end method that applies semantically consistent domain adaptation between synthetic and real traffic scene data. The training of the architecture does not require semantic segmentation maps or pre-trained networks. Our architecture is rather lightweight as it consists of a single encoder, a single decoder, and two discriminators. The lightweight architecture requires less training time and resource allocation, hence reduces energy consumption.  We provide the results of qualitative comparison and quantitative evaluation on the task of semantic segmentation. In both cases, we demonstrate that the proposed method shows significantly improved consistency of the \textit{sim2real} image translation and performance in the underlying task. \href{https://artemsavkin.github.io/secogan}{Project page}.

\section{Related Work}
\subsection{Domain Adaptation}

Machine learning systems assume that the training data and the test data are similar or \textit{i.i.d}. However, in many cases, this assumption does not hold. Domain adaptation aims to minimize the discrepancy between the two domains. This is exceptionally challenging in the unsupervised setup when pairs of the corresponding samples from both domains are not available. Multiple methods fall into the category of unsupervised domain adaptation: entropy minimization \cite{vu2019advent}, curriculum learning \cite{zhang2017curriculum}, generative adversarial networks based approaches \cite{zhu2017unpaired}, and classifier discrepancy \cite{saito2018maximum}. Most acclaimed ones base on adversarial framework (GAN). 

\subsection{Adversarial Domain Adaptation}

The CycleGAN \cite{zhu2017unpaired} is one of the pioneer unsupervised image-to-image translation frameworks. This framework restricts the encoder and decoder by enforcing cycle consistency constraints. Liu et al. \cite{liu2016coupled} adopt a weight sharing mechanism between the layers of generators and discriminators to learn the joint distribution of data. Later they \cite{liu2017unsupervised} proposed a new framework named UNIT which exploits weight sharing and common latent space assumptions. Multi-modal UNIT (MUNIT) \cite{huang2018multimodal} employs the image disentanglement principle into the content and style codes. The image's content code is combined with the cross domain's random style codes to synthesize the diverse outputs in the cross-domain. GAN-based methods produce visually appealing image translations but fail to maintain semantic consistency between source and translated image.

\subsection{Semantic Consistency}

Many works utilize various methods to address the semantic inconsistency problem. A straightforward approach relies on auxiliary information such as semantic maps to track changes in the source and target domains. Hoffman et al. \cite{hoffman2018cycada} introduce a method that preserves semantic consistency by constraining on a cycle consistent task-loss. The task loss tracks the discrepancy between segmentation predictions for the source and the translated images. In another work, Chen et al. \cite{chen2019crdoco} also follow the principle that the same image in different styles should produce identical semantic maps. Using this principle, the authors enforce the adapted model to produce consistent predictions for the same image with different styles. Li et al. \cite{li2019bidirectional} show that the cooperation between image-to-image translation architecture and the segmentation network improves performance.

Integrating a semantic predictor to measure the discrepancy between source and generated images is not the only method to generate semantically consistent image-to-image translation. Li et al. \cite{li2018semantic} introduce the soft gradient-sensitive objective and semantic aware discriminator to retain semantic consistency. Since the alterations in the generated image change the object's boundaries, their method applies the Sobel filter on the image and its corresponding semantic map to track the deviation. According to proponents of the DLOW method \cite{gong2019dlow}, intermediate domains can bridge the gap between the source and the translated image. In their work, multiple target domains are provided for generating multiple intermediate domains. These intermediate domains are used as source data for discriminators in adversarial learning. Apart from the semantic annotation, the depth map could be also incorporated in the discrepancy measuring such as in the work of Chen et al. \cite{chen2019learning}. This approach benefits from the inclusion of depth and semantic annotation as guidelines to transform synthetic images into real images. Another way is the application of domain adaptation to the feature maps of the real and synthetic images. Hong et al. \cite{hong2018conditional} employ a fully convolutional network to transform input images into feature maps, which are then utilized by the discriminator for distinguishing source and target domains. Unlike other methods \cite{Savkin2020}, here the authors do not use pixel distribution or label statistics to perform domain adaptation between synthetic and real images. Our generator performs content disentanglement by employing shared content encoder and shared decoder which operates on learned content feature with fixed style code. Our lightweight generator is constrained on intra-domain and inter-domain reconstruction together with adversarial loss on sample patches. This allows for high-quality style transfer and preserving the semantic consistency. 

\input{figure_networks}
\begin{figure*}[t!]
\begin{adjustbox}{width=1\textwidth}
\begin{tabular}{ccccc}

\subfloat{\includegraphics{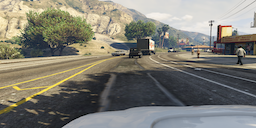}} &
\subfloat{\includegraphics{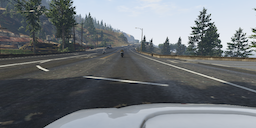}} &
\subfloat{\includegraphics{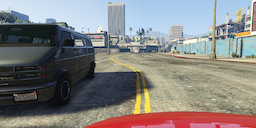}} &
\subfloat{\includegraphics{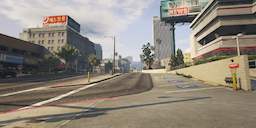}} &
\subfloat{\includegraphics{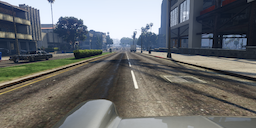}} \\

\subfloat{\includegraphics{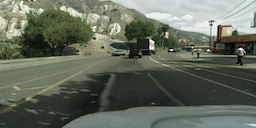}} &
\subfloat{\includegraphics{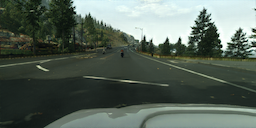}} &
\subfloat{\includegraphics{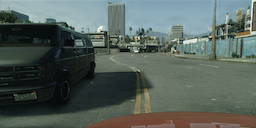}} &
\subfloat{\includegraphics{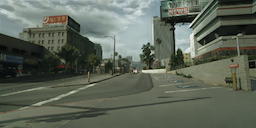}} &
\subfloat{\includegraphics{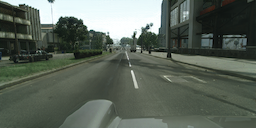}} \\

\subfloat{\includegraphics{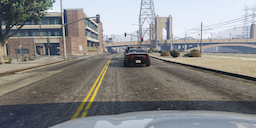}} &
\subfloat{\includegraphics{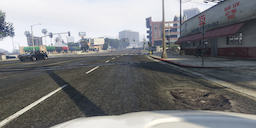}} &
\subfloat{\includegraphics{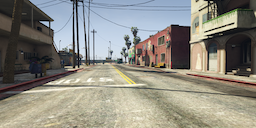}} &
\subfloat{\includegraphics{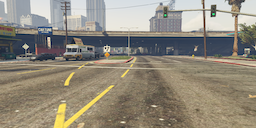}} &
\subfloat{\includegraphics{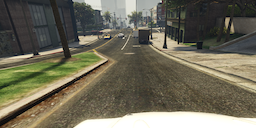}} \\

\subfloat{\includegraphics{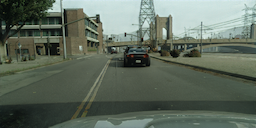}} &
\subfloat{\includegraphics{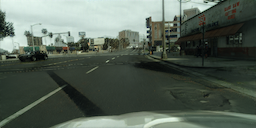}} &
\subfloat{\includegraphics{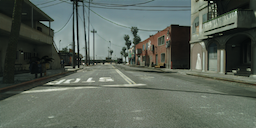}} &
\subfloat{\includegraphics{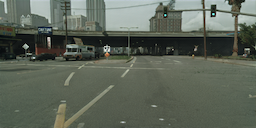}} &
\subfloat{\includegraphics{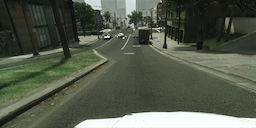}} \\

\end{tabular}
\end{adjustbox}
\caption{Pairs of synthetic source images from PfD \cite{richter2016playing} and translated images generated by proposed method.}
\label{fig:qualitative_results}
\end{figure*}
\section{Approach}

In this work, we follow the assumption that synthetic and real data for urban traffic scenarios reveal common or domain agnostic features describing the content of the scene (e.g. cars, objects, pedestrians) and domain-specific features describing the appearance (e.g. appearance or style). We employ a single network to encode the common features of different domains with randomly sampled style codes for each domain to represent the domain-specific features. Given an extracted content feature vector and a sampled style vector for one of the domains, our single decoder can synthesize an image in the particular domain.  

The overview of our method is demonstrated in Figure~\ref{fig:overview}. In order to perform domain adaptation between two domains, we use encoder $E$, decoder $G$, together with two discriminators $D_i$ for each domain $x_i \in X_i$ ($i=a,b$). As shown in Figure~\ref{fig:overview}, the encoder embeds the input sample $x_i$ into the content code $c_i = E(x_i)$. The content code represents common features between these two domains. Style code for each domain is initiated before the training as $s_i = (\gamma_i, \beta_i)$ and remains constant. It represents the appearance features of each domain. Each style code is a vector of size 256 drawn from a uniform distribution that ranges between 0 and 1. We enforce $c_i$ and $s_i$ to embed content and style features respectively by employing the content-style disentanglement through cross-domain transfer. To generate images cross-domain, the encoder embeds images from the source domain to content code, then normalizes the content code with the target domain's style code via adaptive instance normalization (AdaIN) \cite{huang2017arbitrary}:

\begin{table*}
\caption{Mean IoU values for semantic segmentation prediction by DRN26 trained on synthetic dataset translated to real.}
  \centering
  \begin{adjustbox}{width=\textwidth}
  \small
  \renewcommand{\arraystretch}{2}
    \begin{tabular}{c|c|c|cc|ccccccccccccccccccccc|}
      \hline
    \rotatebox[origin=c]{90}{Method} & \rotatebox[origin=c]{90}{Train} & 
    \rotatebox[origin=c]{90}{Eval} & \rotatebox[origin=c]{90}{Accuracy} & \rotatebox[origin=c]{90}{mean IoU} & \rotatebox[origin=c]{90}{road} & \rotatebox[origin=c]{90}{sidewalk} & \rotatebox[origin=c]{90}{building} & \rotatebox[origin=c]{90}{wall} & \rotatebox[origin=c]{90}{fence} & \rotatebox[origin=c]{90}{pole} & \rotatebox[origin=c]{90}{ traffic light } & \rotatebox[origin=c]{90}{traffic sign} & \rotatebox[origin=c]{90}{vegetation} & \rotatebox[origin=c]{90}{terrain} & \rotatebox[origin=c]{90}{sky} & \rotatebox[origin=c]{90}{person} & \rotatebox[origin=c]{90}{rider} & \rotatebox[origin=c]{90}{car} & \rotatebox[origin=c]{90}{truck} & \rotatebox[origin=c]{90}{bus} & \rotatebox[origin=c]{90}{train} & \rotatebox[origin=c]{90}{motorbike} & \rotatebox[origin=c]{90}{bicycle}\\
      \hline
      None & CS & CS & 94.3 & 67.4 & 97.3 & 79.8 & 88.6 & 32.5 & 48.2 & 46.3 & 63.6 & 73.3 & 89.0 & 58.9 & 93.0 & 78.2 & 55.2 & 92.2 & 45.0 & 67.3 & 39.6 & 49.9 & 73.6  \\
      \hline
      None & GTA & CS & 62.5 & 21.7 & 42.7 & 26.3 & 51.7 & 5.5 & 6.8 & 13.8 & 23.6 & 6.9 & 75.5 & 11.5 & 36.8 & 49.3 & 0.9 & 46.7 & 3.4 & 5.0 & 0.0 & 5.0 & 1.4 \\
      CycleGAN & GTA \(\longrightarrow\) CS & CS & 82.5 & 32.4 & 81.8 & 34.7 & 73.5 & 22.5 & 8.7 & 25.4 & 21.1 & 13.5 & 71.5 & 26.5 & 41.7 & 50.1 & 7.3 & 78.5 & 20.5 & 19.5 & 0.0 & 12.5 & 6.9 \\
      DRIT & GTA \(\longrightarrow\) CS & CS & 78.9 & 27.4 & 78.8 & 26.1 & 68.7 & 13.1 & 10.2 & 18.8 & 11.4 & 17.9 & 56.8 & 4.8 & 34.1 & 45.8 & 9.4 & 73.3 & 13.4 & 14.4 & 10.5 & 7.7 & 5.3 \\
      MUNIT & GTA \(\longrightarrow\) CS & CS & 85.8 & 37.4 & \textbf{85.9} & \textbf{35.9} & 79 & 26.1 & \textbf{18.4} & 31.2 & 23.8 & 17.1 & 74.4 & 17.4 & 55.8 & 52.3 & \textbf{17.3} & 82.5 & 21.2 & 22.8 & \textbf{12.0} & \textbf{20.7} & \textbf{16.7}\\
      CUT & GTA \(\longrightarrow\) CS & CS & 82.6 & 34.0 & 78.7 & 26.7 & 73.3 & 17.4 & 16.9 & 22.2 & 25.7 & 16.2 & 73.4 & 20.8 & 64.8 & 54.8 & 13 & 79.3 & 20.8 & 22.2 & 0 & 11.6 & 8.6 \\
      Ours & GTA \(\longrightarrow\) CS & CS & \textbf{88.7} & \textbf{39.0} & 82.0 & 28.1 & \textbf{80.7} & \textbf{30.8} & 15.0 & \textbf{32.0} & \textbf{34.4} & \textbf{23.6} & \textbf{80.1} & \textbf{34.3} & \textbf{76.9} & \textbf{55.7} & 10.2 & \textbf{82.9} & \textbf{25.5} & \textbf{26.5} & 1.6 & 11.9 & 8.8 \\
     \hline
    \end{tabular}
    \end{adjustbox}
    \label{tab:ToRef}
\label{tab:results}
\end{table*}

\begin{equation}
    \text{AdaIN}(z,\gamma,\beta) = \gamma \left(\frac{z - \mu(z)}{\sigma(z)} \right) + \beta
\end{equation}

where z is the activation of the encoder output, $\gamma$ and $\beta$ are the style code parameters of the target domain. We also introduce several constraints to accurately reconstruct or translate the image to the target domain with the decoder. 

The first constraint is enabled by accurate reconstruction of the input image. The encoder embeds the input image into content code; then, the decoder combines one domain's embedding with its style codes. Here, the network minimizes the reconstruction loss, which is defined as follows: 

\begin{equation}
    \mathcal{L}_{rec}^{aa} (E,G) = \mathbb{E}_{{x_a} \sim X_a} \|G(E(x_a), s_a) - x_a\|_1
\end{equation}

To enforce consistency further, we also employ cycle reconstruction \cite{zhu2017unpaired}. After translating the source input $x_a$ to the target domain, the translated image $x_{ab}$ is then translated cross-domain back again resulting in $x_{aba}$. Therefore, the network should be able to reconstruct input after translating it to the target and back to the source domain. To do that we utilize the cycle reconstruction loss defined as follows:    

\begin{equation}
    \mathcal{L}_{cyc}^{aba} (E,G) = \mathbb{E}_{{x_a} \sim X_a} \|G( E( x_{ab} ), s_{a}) - x_a\|_1
\end{equation}

where $x_{ab}$ denotes the translated image from domain $a$ to domain $b$. Lastly, adversarial learning is applied to match the distribution of translated images to the target domain distribution. The generated images in the cross-domain should be indistinguishable from real images in the target domain. In our work, the discriminator only receives the random patches $p$ of source and translated images \cite{park2020swapping}. Computation of reconstruction, cycle reconstruction and adversarial losses are depicted Figure~\ref{fig:overview}. We utilize \textit{adversarial loss} to match the data distribution of translated images to the cross domain data distribution and adopt visual characteristics of target domain: 

\begin{equation}
    \begin{split}
    \mathcal{L}_{adv}^{a} (E,G,D_a) & = \mathbb{E}_{{x_a} \sim X_a} log D_{a}(p(x_{a})) 
    \\ & + \mathbb{E}_{{x_b} \sim X_b} log(1-D_{a}(p(x_{ba})))
    \end{split}
\end{equation} 

where $p$ selects the random patches from the image and $x_{ba}$ denotes the translated image from domain $b$ to domain $a$. We note the other loss terms  $\mathcal{L}_{rec}^{bb}$, $\mathcal{L}_{cyc}^{aba}$, and $\mathcal{L}_{adv}^{b}$ are defined similarly w.r.t domains. The overall loss function for the generator is given in the following equation.


\begin{equation}
    \begin{split}
    \underset{E,G}{\min} \: \underset{D_a,D_b}{\max} 
    \mathcal{L}_{} (E,G,D_a,D_b) &= \lambda_1 (\mathcal{L}_{rec}^{aa} + \mathcal{L}_{rec}^{bb}) \\ &+ \lambda_2 (\mathcal{L}_{cyc}^{aba} + \mathcal{L}_{cyc}^{bab}) \\ &+ \lambda_3 (\mathcal{L}_{adv}^{a} + \mathcal{L}_{adv}^{b})
    \end{split}
\label{eqn:GeneratorLoss}
\end{equation}

where $\lambda_1$, $\lambda_2$, and $\lambda_3$ define the contribution of each component.

\begin{figure*}

\hspace{10mm} \textbf{PfD} \hspace{16mm} \textbf{CycleGAN} \hspace{14mm} \textbf{MUNIT} \hspace{17mm} \textbf{DRIT} \hspace{20mm} \textbf{CUT} \hspace{20mm} \textbf{Ours}

\begin{adjustbox}{width=\textwidth}
\begin{tabular}{c c c c c c}

\subfloat{\includegraphics{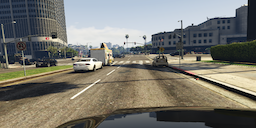}} &
\subfloat{\includegraphics{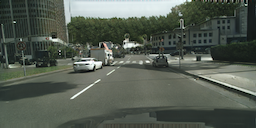}} &
\subfloat{\includegraphics{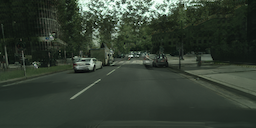}} &
\subfloat{\includegraphics{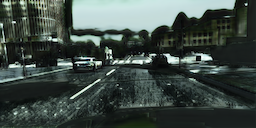}} &
\subfloat{\includegraphics{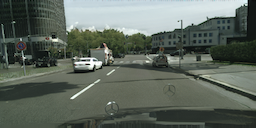}} &
\subfloat{\includegraphics{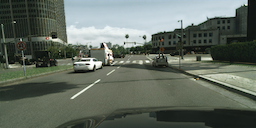}}\\

\subfloat{\includegraphics{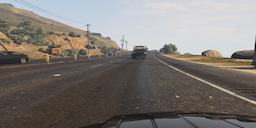}} &
\subfloat{\includegraphics{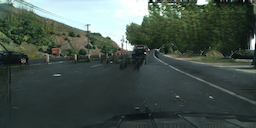}} &
\subfloat{\includegraphics{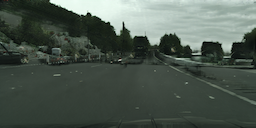}} &
\subfloat{\includegraphics{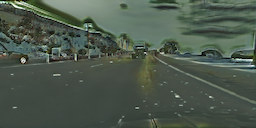}} &
\subfloat{\includegraphics{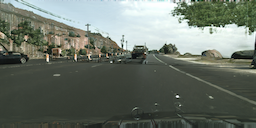}} &
\subfloat{\includegraphics{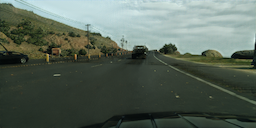}}\\

\subfloat{\includegraphics{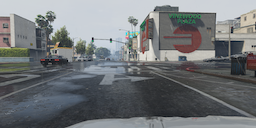}} &
\subfloat{\includegraphics{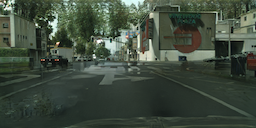}} &
\subfloat{\includegraphics{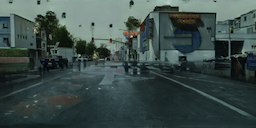}} &
\subfloat{\includegraphics{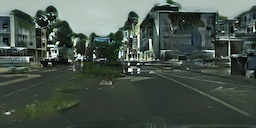}} &
\subfloat{\includegraphics{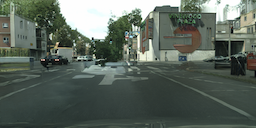}} &
\subfloat{\includegraphics{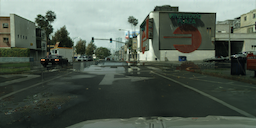}}\\

\subfloat{\includegraphics{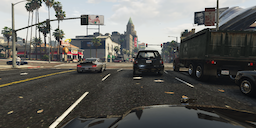}} &
\subfloat{\includegraphics{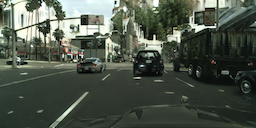}} &
\subfloat{\includegraphics{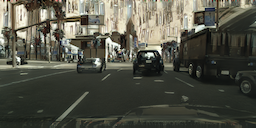}} &
\subfloat{\includegraphics{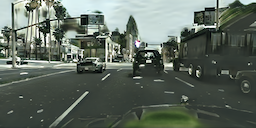}} &
\subfloat{\includegraphics{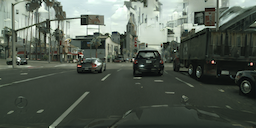}} &
\subfloat{\includegraphics{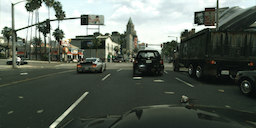}}\\

\subfloat{\includegraphics{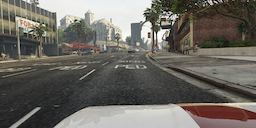}} &
\subfloat{\includegraphics{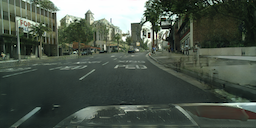}} &
\subfloat{\includegraphics{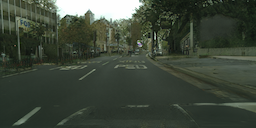}} &
\subfloat{\includegraphics{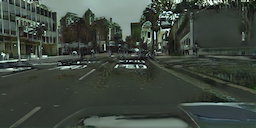}} &
\subfloat{\includegraphics{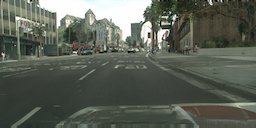}} &
\subfloat{\includegraphics{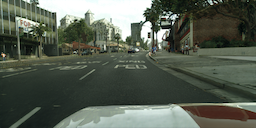}}\\

\subfloat{\includegraphics{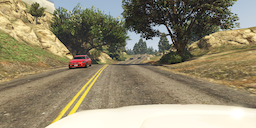}} &
\subfloat{\includegraphics{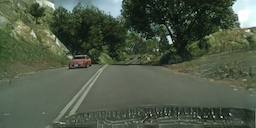}} &
\subfloat{\includegraphics{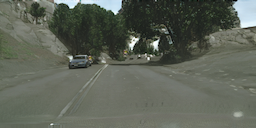}} &
\subfloat{\includegraphics{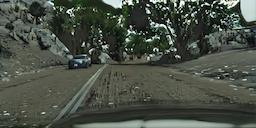}} &
\subfloat{\includegraphics{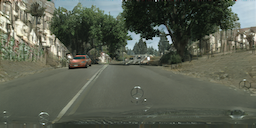}} &
\subfloat{\includegraphics{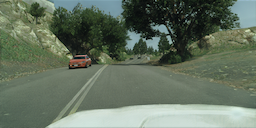}}\\

\end{tabular}
\end{adjustbox}

\caption{Semantic consistency of the images translated by proposed method compared to baselines.}
\label{fig:comparison}
\end{figure*}
\section{Experiments} 
\subsection{Network Architecture}

Our network consists of one generator (encoder and decoder) and two discriminators. Encoder and decoder architectures are based on MUNIT \cite{huang2018multimodal} implementation and discriminator architecture is based on PatchGAN \cite{isola2017image} implementation. Each discriminator obtains the random patches of translated and source images. The random patch size varies from 1/8 to 1/4 of the full image size on each side. A detailed overview of the encoder-decoder network is given in the Figure~\ref{fig:networks}.

The proposed adaptation method is unsupervised as it does not require correspondence between source and target samples. Furthermore, it needs neither pre-trained segmentation networks nor semantic or depth maps to constraint the learning and achieve semantic consistency.

We set $\lambda_1$,  $\lambda_2$, and $\lambda_3$ parameters in Equation~\ref{eqn:GeneratorLoss} to $10$, $10$, and $1$ respectively. We also use mini-batch stochastic gradient descent with Adam optimizer \cite{kingma2014adam}. Beta coefficients of Adam are set to $0.5$ and $0.999$, respectively. The training of the network continues for 200 epochs with 2975 randomly sampled PfD \cite{richter2016playing} images (GTA-3k) and 2975 Cityscapes \cite{cordts2016cityscapes} images (CS-3k) for training.

\subsection{Datasets}

We use in experiments two large-scale datasets for urban traffic scenes: real Cityscapes \cite{cordts2016cityscapes} and synthetic PfD \cite{richter2016playing}.

\textbf{Cityscapes} includes images of traffic scenes recorded in 50 different cities in Germany and France \cite{cordts2016cityscapes}. The dataset provides 5000 fine-annotated image pairs, of which 2975 are publicly available for training. The dataset contains 19 annotated classes: road, sidewalk, building, wall, fence, pole, traffic light, traffic sign, vegetation, terrain, sky, person, rider, car, truck, bus, train, motorcycle, and bicycle. The images in the Cityscapes have a resolution of $2048\times1024$ pixel.

\textbf{PfD} contains 24966 photo-realistically rendered images from the video game GTA \cite{richter2016playing}. The images have a resolution of 1914$\times$512 pixels. The number of semantic classes is 19 and they are aligned with the classes in Cityscapes. In our experiments, we resize both Cityscapes and PfD due to the memory and time constraints to 1024$\times$512 pixels.

\subsection{Comparison} 

As our method does not rely on any auxiliary methods such as pre-trained segmentation models or semantic maps to enhance the image-to-image translation, we compare it to baseline techniques that also perform end-to-end translation in unsupervised way. For that matter, we pick CycleGAN \cite{zhu2017unpaired}, MUNIT \cite{huang2018multimodal}, DRIT \cite{lee2018diverse}, and CUT \cite{park2020contrastive}. For the MUNIT network we disabled the domain-invariant perceptual loss, since it leverages a pre-trained recognition model.

Similar to our network baseline models are trained on PfD-3k and CS-3k with the sample size of $1024\times512$ pixels. In the testing phase, the full PfD dataset (24966 images) is translated to the Cityscapes domain by each method picked for comparison. These translated PfD datasets are then used for the training of semantic segmentation networks. The performance scores of the segmentation network on the Cityscapes \textit{val} provide the basis for quantitative evaluation. A better segmentation score indicates a better translation. 

\subsection{Qualitative Results} 

The results of the translation of our network are depicted in Figure~\ref{fig:qualitative_results}. Here we want to highlight such aspects of translation as quality of style transfer (road texture, color of lane marking) and preserved semantics of translated image. Qualitative comparison to the baseline methods is provided in Figure~\ref{fig:comparison}. Here each row demonstrates the original synthetic source image together with the images translated by respective baseline method. It is worth mentioning that CycleGAN translates the car as a part of the road and also introduces vegetation-alike patches in the sky regions. Comparison with the MUNIT shows that MUNIT produces image-to-image translations in different styles due to its style sampling part. In the first example, MUNIT covers the whole sky with vegetation and struggles in keeping the semantic content unchanged. Similar to CycleGAN, MUNIT also translates the ego car as a part of the road. DRIT follows the design principles of MUNIT where both architectures disentangle input into shared latent vectors and domain-specific attribute vectors. DRIT struggles from inconsistency as well. CUT demonstrates that image-to-image translation can be performed by increasing the mutual information between the patches from the same regions of the source and translated images. The architecture uses the networks from CycleGAN and implementation shares the same problems with CycleGAN in translation and adds other artifacts such as in-painting of the car logo which is typical for Cityscapes dataset. Our method in turn shows consistency for all classes, especially \textbf{vegetation, sky, road} and also ego car.

\subsection{Quantitative Results}

In this part, we trained DRN-C-26 \cite{yu2017dilated} network on the dataset generated by translation of GTA images to Cityscapes. The performance of the segmentation algorithm is measured on the Cityscapes \textit{val}. 

The first row of Table~\ref{tab:results} shows the performance of the segmentation algorithm when it is trained the Cityscapes \textit{train} and tested with the Cityscapes \textit{val}. This represents the oracle and the upper bound. The second row shows the result when the segmentation algorithm is trained with the PfD dataset and tested also Cityscapes \textit{val}. Other rows show translation performance of the CycleGAN, DRIT, MUNIT, and CUT implementations. Here one can see that our method improves segmentation accuracy by $+2,9$ points achieving $88,7$ and mean \textit{intersection over union} measure by $+1,6$ achieving $39,0$. Our method also shows improved performance for multiple classes, such as \textit{building, vegetation, sky, car} etc.

\subsection{Content Space}

\begin{figure}[!t]
\centering
\includegraphics[width=1\columnwidth]{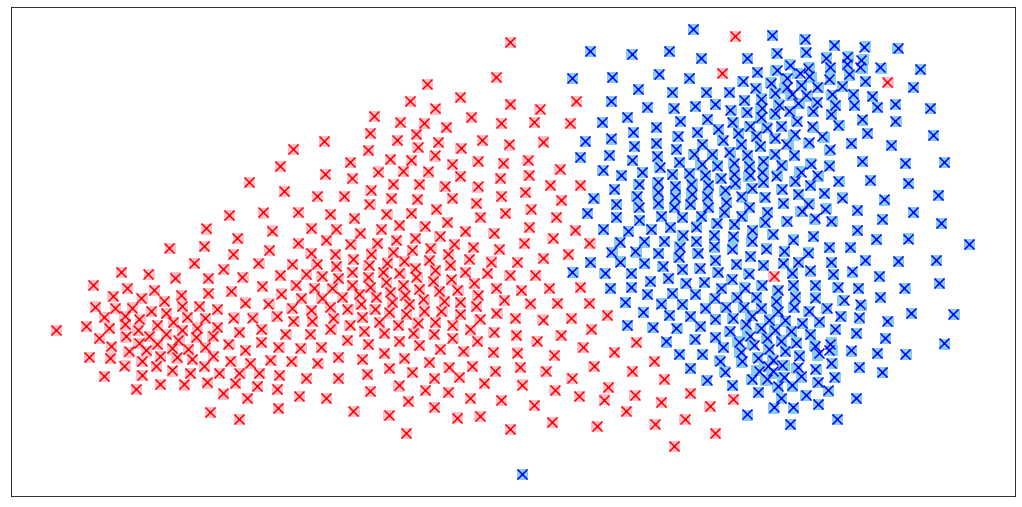}


\caption{t-SNE\cite{Roweis2002} projections derived for randomly picked codes $c_a$ (red $\square$), $c_{ab}$ (red $\times$), $c_b$ (blue $\square$), $c_{ba}$ (blue $\times$).}
\label{fig:tsne}
\end{figure}

Furthermore, we analyse the ability of our method to extract meaningful content vectors by applying \textit{t-NSE} \cite{Roweis2002} algorithm to the content vectors extracted during the 200 epoch of the training. Figure~\ref{fig:tsne} shows that content codes extracted from source images match with the codes extracted from cross-domain translated images. Thus, $c^i_a$ match with $c^i_{ab}$ and $c^j_b$ match with $c^j_{ba}$. This confirms the disentanglement from style codes as we do not apply any constraints on content vector directly but rather learn them implicitly via reconstruction and adversarial loss.

\begin{figure}[!t]
\centering
\includegraphics[width=1\columnwidth]{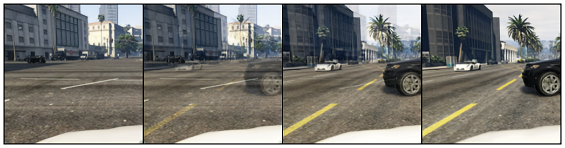}
\includegraphics[width=1\columnwidth]{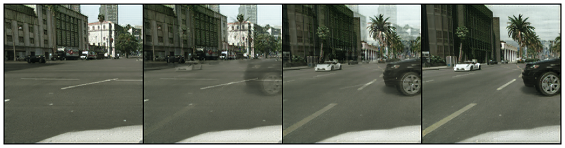}


\caption{Interpolation results for two content vectors $c^1_a$ (left) and $c^2_a$ (right) with source (top) and target styles (bottom).}
\label{fig:interpolation}
\end{figure}
\section{Conclusion}\label{sec:conclusion}

In this paper, we propose an unsupervised method for semantically consistent synthetic-to-real domain adaptation of urban traffic scenes. We compared our method with state-of-the-art networks in synthetic-to-real image translation. Visual and quantitative comparisons on synthetic-to-real image translation show that our architecture improves the visual quality of translated images as well as the performance of deep semantic segmentation network trained on the translated images. 

\section{Acknowledgement}
The research leading to these results is funded by the German Federal Ministry for Economic Affairs and Energy within the project “KI Absicherung – Safe AI for Automated Driving". The authors would like to thank the consortium for the successful cooperation. 

\newpage
{
\bibliographystyle{ieee}
\bibliography{references}
}

\end{document}